\documentclass[conference]{IEEEtran}
\IEEEoverridecommandlockouts
\usepackage[noadjust]{cite}
\usepackage{amsmath,amssymb,amsfonts}
\usepackage{algorithmic}
\usepackage{graphicx}
\usepackage{textcomp}
\usepackage{makecell}
\usepackage{float}
\usepackage{url}          
\usepackage{breakurl}     
\usepackage{xcolor}
\usepackage{tabularx}
\usepackage{subcaption}
\let\svthefootnote\thefootnote
\newcommand\freefootnote[1]{%
  \let\thefootnote\relax%
  \footnotetext{#1}%
  \let\thefootnote\svthefootnote%
}
\def\BibTeX{{\rm B\kern-.05em{\sc i\kern-.025em b}\kern-.08em
    T\kern-.1667em\lower.7ex\hbox{E}\kern-.125emX}}
\begin{document}

\title{Enhancing SDG-Text Classification with Combinatorial Fusion Analysis and Generative AI}


\author{
    \IEEEauthorblockN{
        Jingyan Xu\IEEEauthorrefmark{1},
        Marcelo L. LaFleur\IEEEauthorrefmark{2},  
        Christina Schweikert\IEEEauthorrefmark{3}, 
        and D. Frank Hsu\IEEEauthorrefmark{1},~\IEEEmembership{Senior Member, IEEE}\\
    }
    \IEEEauthorblockA{
        \IEEEauthorrefmark{1}Laboratory of Informatics and Data Mining, \\
        Department of Computer and Information Science, Fordham University, New York, NY, USA\\
    }
    \IEEEauthorblockA{
        \IEEEauthorrefmark{2}Dept. of Economic and Social Affairs, United Nations, New York, NY, USA\\
    }
    \IEEEauthorblockA{
    \IEEEauthorrefmark{3}Division of Computer Science, Mathematics and Science, St. John’s University, Queens, NY, USA\\
    }
}

\maketitle

\begin{abstract}
(Natural Language Processing) NLP techniques such as text classification and topic discovery are very useful in many application areas including information retrieval, knowledge discovery, policy formulation, and decision-making. However, it remains a challenging problem in cases where the categories are unavailable, difficult to differentiate, or are interrelated. Social analysis with human context is an area that can benefit from text classification, as it relies substantially on text data. The focus of this paper is to enhance the classification of text according to the UN's Sustainable Development Goals (SDGs) by collecting and combining intelligence from multiple models. Combinatorial Fusion Analysis (CFA), a system fusion paradigm using a rank-score characteristic (RSC) function and cognitive diversity (CD), has been used to enhance classifier methods by combining a set of relatively good and mutually diverse classification models. We use a generative AI model to generate synthetic data for model training and then apply CFA to this classification task. The CFA technique achieves 96.73\% performance, outperforming the best individual model. We compare the outcomes with those obtained from human domain experts. It is demonstrated that combining intelligence from multiple ML/AI models using CFA and getting input from human experts can, not only complement, but also enhance each other. 
\end{abstract}


\section{Introduction}

\freefootnote{\textcopyright 2025 IEEE. Personal use of this material is permitted. Permission from IEEE is required for all other uses.}

Text classification plays a crucial role in many applications including policy formulation and decision making, as it extracts valuable insights from large volumes of unstructured text data \cite{jin_natural_2023}. By automatically identifying potential themes and patterns within these texts, policy-makers can efficiently navigate through complex documents and find relevant information pertaining to specific policy areas. Text classification can also help monitor progress and assess the effectiveness of policies over time. Tracking changes in topic distributions allows policymakers to observe trends, measure the success of implemented policies and identify areas for improvement or adjustment. It provides the necessary transparency and ensures alignment with proposed objectives. 

\begin{figure}
    \centering
    \includegraphics[width=0.85\linewidth]{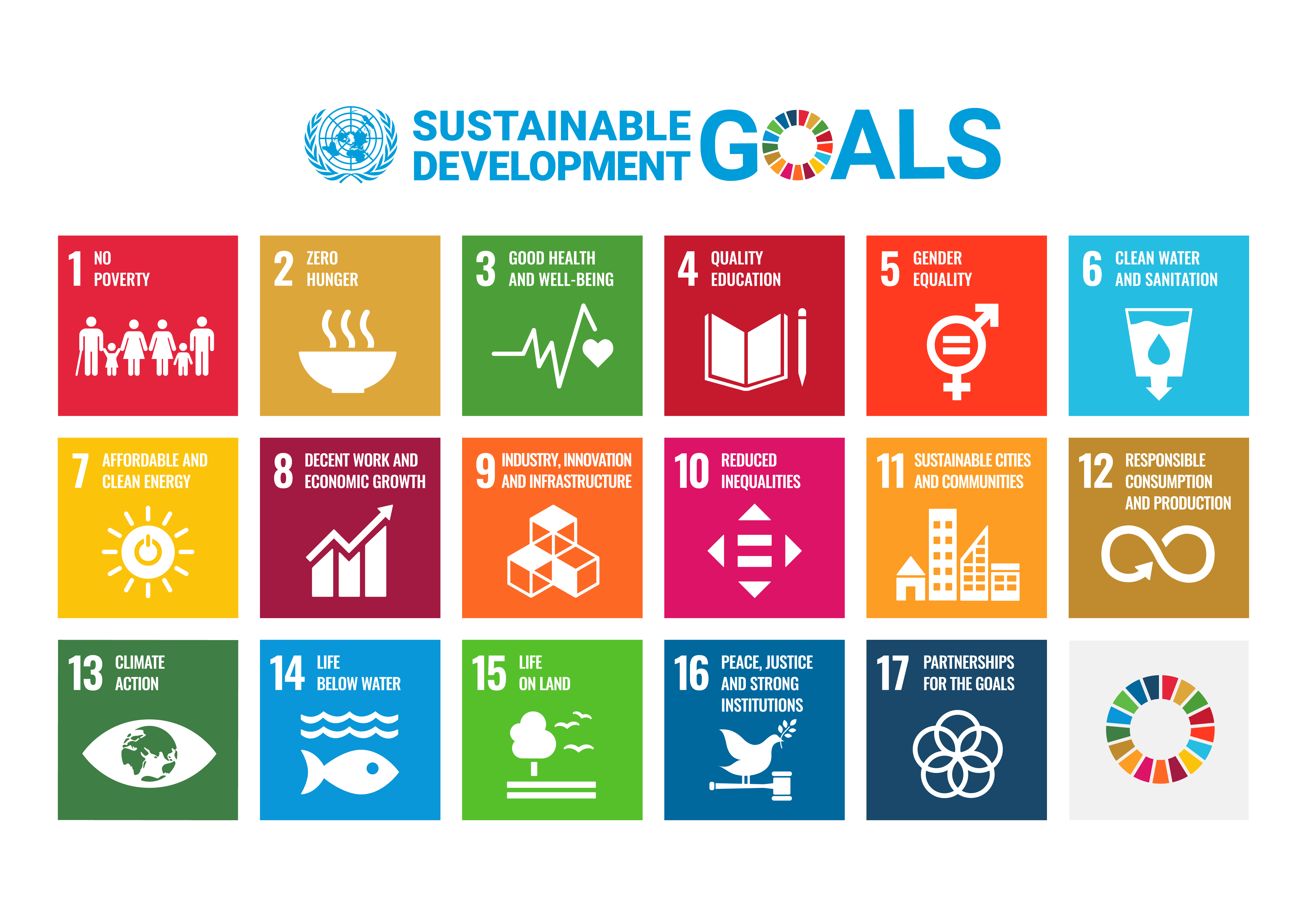}
    \caption{SDGs of the United Nations \cite{UN_picture}}
    \label{fig:sdg figure}
\end{figure}

The United Nations (UN) established 17 Sustainable Development Goals (SDGs) aimed at fostering prosperity for the Earth and humanity \cite{biermann_global_nodate}, shown in Fig. \ref{fig:sdg figure}. The SDGs represent a culmination of more than a decade's worth of collaborative efforts among participating nations. They are in essence a continuation of the work of 8 Millennium Development Goals (MDGs) that ran from 2000 to 2015. The MDGs were instrumental in lifting nearly one billion individuals out of extreme poverty, addressing hunger, and increasing access to education for girls. For example, MDG goal 7 made significant contributions to environmental protection by nearly eliminating global consumption of ozone-depleting substances, promoting reforestation initiatives, and expanding conservation efforts worldwide. The SDGs, with their ambitious development agenda, aim to maintain the progress initiated by the MDGs. 

Classifying texts produced by the UN helps to promote these goals, but SDGs are concepts that are interconnected, which presents unique challenges for machine learning classification \cite{lafleur_using_nodate}. Each SDG covers a broad spectrum of themes, with progress in one often reliant on progress in others. This interconnectedness introduces subjectivity and inconsistency in labeling which are difficult to avoid. While it may be feasible to assign an SDG to a single sentence, accurately quantifying how a longer text addresses these interconnected goals requires subjective judgement by human experts, making consistent classification challenging. 

The complexity of these United Nations documents make it possible to have multiple ground truths, a counter-intuitive phenomenon that will be shown in our testing dataset. It's a problem that can be dealt with by assigning primary, secondary, tertiary, etc. labels to these documents, but there needs to be a reliable way to generate sufficiently large number of experts and obtain the most appropriate labels.In this paper, we apply the methodology of model fusion to determine these labels.

Moreover, another key challenge lies in the scarcity of labeled data for training and testing models \cite{lafleur_using_nodate}. It is difficult to acquire the volume of data needed to represent the full scope of all SDGs. The manual curation of text by human experts according to 17 SDGs is labor-intensive and difficult to do en masse. Datasets used for machine learning algorithms vary in size and are huge in volume. The costs involved in manually labeling this volume of data are substantial. While machine learning methods in general can ease the problem by automating the process, training machine learning classifiers usually require large training datasets that can provide the algorithm with examples of the "right answers". The fact that there does not exist a dedicated dataset for SDG classification that's sufficiently large highlights the need to leverage generative AI to augment the existing training set. A study has shown that manually labeling 3000 samples for an SST-2 task, a movie review sentiment analysis dataset, would roughly cost 221 to 300 US dollars and around 1000 minutes \cite{ding-etal-2023-gpt}.

This paper applies a synthetic data generative AI model, namely ChatGPT, and Combinatorial Fusion Analysis (CFA) \cite{hsu_combinatorial_nodate, hsu_kristal_2024}, a system fusion paradigm, to address these unique challenges. Section 2 provides a literature review. Section 3 details the methods used including CFA and the five base classification models. Section 4 includes the experiment's methodology flow with an overview of the datasets. Section 5 discusses CFA results obtained by various CFA combinations and compares those results with those of base models as well as human experts. In particular, it is demonstrated that CFA (as machine intelligence) and human experts (as natural intelligence) can complement and enhance each other. Finally, Section 6 concludes the paper with some remarks on advantages and limitations.

\section{Literature Review}
Text (or document) classification involves using machine learning algorithms to assign class labels to instances \cite{bishop_pattern_2006}. In supervised learning, this process requires a training dataset with numerous examples of inputs and corresponding outputs for learning. A model learns to map input text data to specific class labels from this dataset. As such, the training dataset must adequately represent the problem and include a sufficient number of examples for each class. On the other hand, unsupervised learning methods like clustering are used to handle unlabeled data. Association algorithms are also useful for establishing rules of relationships, such as identifying that customers who purchase product X are also likely to purchases product Y. 

With the development of large language models, data augmentation for Natural Language Processing (NLP) became increasingly popular in recent years. A previous study used ChatGPT to generate synthetic text that are representative of the SDG concepts \cite{lafleur_using_nodate}. The result indicates a significant advantage as this technique is relatively low cost compared to manually labeling documents, and that the documents generated can mirror the language used in the vast amount of documents that discuss the SDGs. 

There are some existing classifiers that are designed to classify data items according to SDGs. Among them, SDG Mapper, developed by the European Union in 2017, and LinkedSDG, developed by the United Nations Department of Economic and Social Affairs, are used in the current study. In addition, we also use another classifier that's developed in 2019 by an SDG domain expert \cite{lafleur_art_2019}. A description of these models will be included in the latter section. 

As transformer models became more popular, additional SDG tools were developed in 2022 by leveraging this model, such as SDG Meter \cite{guisiano_sdg-meter_nodate} and EUR-SDG-Mapper \cite{jelicic_eur-sdg-mapper_nodate} which were built using Bidirectional Encoder Representations from Transformers (BERT) \cite{devlin_bert_nodate}, a neural network-based language model. BERT operates on an architecture based on the mechanism of self-attention, which assigns varying degrees of importance to different segments of input data, thereby providing context for any position within the input sequence. It works by performing a small and constant number of steps. In each of the steps, it applies the attention mechanism to understand relationships between all words in a sentence. SDG Prospector was constructed using DistilRoBERTa, a lighter version of the RoBERTa-based model developed by Meta \cite{jacouton_proof_2022}. This model, a sentence-transformers model, maps text to a 768 dimensional dense vector space and is applicable for tasks such as clustering or semantic search. Just like BERT, it's also a neural network model pretrained on millions of texts with a very diverse vocabulary and complex English grammar rules and conjugations. 

The works mentioned above are some of the tools that exist to analyze SDGs. Most studies on SDG classification have typically relied on a single model. Our current study adopts a fusion approach by combining intelligence from multiple models. In machine learning, there are techniques like bagging and boosting in ensemble methods that are designed to enhance model accuracy by combining multiple models rather than relying on a single one \cite{zhou_ensemble_2012}. CFA is an approach that provides methods to combine models and can significantly improve overall performance \cite{hsu_combinatorial_nodate, hsu_kristal_2024, hurley_multi-layer_2021}. 

Hsu, LaFleur, and Orazbek's  work \cite{hsu_improving_2022} established the precedence of using CFA to improve SDG classification results. Their work used two base models, SDG Classy \cite{lafleur_art_2019} and LinkedSDG \cite{noauthor_linkedsdg_nodate}, which are also used in this study. It examined the classification results from combining the models using score and rank combinations, although weighted combinations weren't used. It evaluated the relationship between cognitive diversity and precision by dividing cognitive diversity into three groups: low, middle, and high. It demonstrated that combinations can help improve classification precision only if the individual models have high performance and are diverse. 

\section{Methods}
The methodology of Combinatorial Fusion Analysis (CFA) and how it utilizes cognitive diversity and various methods of combination to improve performance are discussed in this section. In addition, the five base models used to classify documents w.r.t. to the SDGs are also described.

\subsection{Combinatorial Fusion Analysis}

CFA addresses an important question for machine learning practitioners, which is whether model fusion is necessary to improve performance. A crucial part of training ML algorithms is hyperparameter tuning, where ML engineers choose different sets of hyperparameters, each affecting the model's performance differently. Traditionally, engineers would train multiple models with various hyperparameter configurations and then select the best-performing one as the "final" model and discard the rest. This process, however, is inefficient as it consumes much computational resources. 

It is shown through existing research that using a model fusion approach, like CFA, can yield better performance compared to a single optimized model as it can reduce overfitting, increase robustness, and leverage the strengths and weaknesses of different models. When we are uncertain about whether there exists an individual model that can consistently outperform the others, we should consider a combination of models that works reasonably well, albeit not optimal. CFA has shown improved performance on several different domain applications including, but not limited to, drug discovery \cite{jiang_enhancing_2023}, information and network security \cite{10288981}, portfolio management \cite{wang_improving_2019}, and material science \cite{tang_improving_2021}. 

Let D=$\{d_1, d_2, ..., d_n\}$ be a set of $n$ documents, instances, or subjects. Under CFA, a scoring system $A$ produces a score value, $s_A(d_i)$, where $d_i$ is in $D$, and the score values are from the set of real numbers $\mathbb{R}$. This set of score values is then normalized to the range of [0, 1]. A rank function $r_A$ with the range of natural numbers $\mathbb{N}$ is generated for each score in the set, with the lowest rank corresponding to the highest score. The rank-score function, $f_A$, depicts the function from rank $i$ in the rank space, $\mathbb{N}$, to the score $f_A(i)$ in Euclidean space such that $f_A(i)=s_A(r_A^{-1}(i))=(s_A\circ r_A^{-1})(i), i\in \mathbb{N}$ \cite{hsu_combinatorial_nodate, hsu_rank-score_2010, hsu_kristal_2024, hurley_multi-layer_2021}.

\begin{figure}
    \centering
    \includegraphics[width=0.80\linewidth]{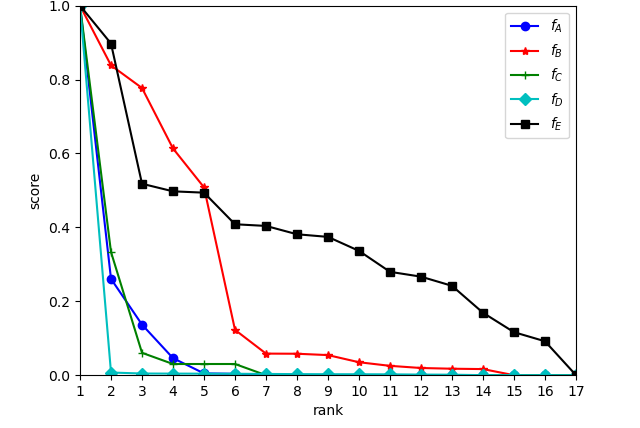}
    \caption{RSC function graph w.r.t. models A, B, C, D, and E for document indicator 2-1-1}
    \label{fig:RSC}
\end{figure}

The concept of the rank-score characteristic (RSC) function was defined by Hsu, Shapiro, and Taksa \cite{hsu_methods_2002}. An example of an RSC function graph that represents the scoring behavior of the base models for one of the documents in this study is shown in Figure \ref{fig:RSC}. It's been used to characterize the scoring behaviors of different models and to measure dissimilarity/diversity between scoring systems in the dual space of Euclidean space and rank space. More specifically, \textbf{cognitive diversity (CD)} between scoring systems A and B, CD(A, B), quantifies the dissimilarity by computing the area between $f_A$ and $f_B$ using the following formula \cite{hsu_kristal_hao_2019}: 

\vspace{-1em}

\begin{equation}
CD(A, B)=d(f_A, f_B)=\sqrt{\frac{\sum^{n}_{i=1}(f_A(i)-f_B(i))^2}{n-2}}
\end{equation}

\vspace{-0.5em}

\noindent where $n$ represents the number of data items in $D$.

Suppose we have a set of $t$ scoring systems, namely $A_1, A_2, ..., A_t$, the \textbf{diversity strength} of the scoring system $A_j$, denoted as $ds(A_j)$, is defined as the arithmetic average of cognitive diversity between $A_j$ and other scoring systems $A_k, k\neq j$\cite{ jiang_enhancing_2023}. This is expressed as: 

\vspace{-0.7em}

\begin{equation}
ds(A_j)=\frac{\sum_{k\neq j}d(A_j, A_k)}{t-1}
\end{equation}

\vspace{-0.7em}

\noindent where $d(A_j, A_k)$ is the cognitive diversity between $A_j$ and $A_k$ \cite{hsu_kristal_hao_2019}. 

When combining the $t$ scoring systems, different types of combinations can be considered: \textbf{average combination}, \textbf{weighted combination by diversity strength}, and \textbf{weighted combination by performance} \cite{hsu_combinatorial_nodate, hsu_kristal_2024, jiang_enhancing_2023}. In addition, each of the three types of combinations can be used for both score and rank combination.

For average score combination (ASC) or average rank combination (ARC), the score function for the score combination $s_{SC}$ and of the rank combination $s_{RC}$ are as follows \cite{hsu_kristal_2024, jiang_enhancing_2023}: 

\vspace{-1em}

\begin{equation}
s_{ASC}(d_i)=\frac{\sum^t_{j=1}s_{A_j}(d_i)}{t}
\end{equation}

\vspace{-1em}

\begin{equation}
s_{ARC}(d_i)=\frac{\sum^t_{j=1}r_{A_j}(d_i)}{t}
\end{equation}

\vspace{-0.7em}

\noindent where ASC and ARC are average score combination and average rank combination, respectively. 

In the case of weighted combination, whether by diversity strength or performance, the score function for \textbf{weighted score combination} and \textbf{weighted rank combination} are expressed as \cite{hsu_kristal_2024, jiang_enhancing_2023}: 

\vspace{-1em}

\begin{equation}
s_{WSC}(d_i)=\frac{\sum^t_{j=1}w_j\times s_{A_j}(d_i)}{\sum^t_{j=1}w_j}
\end{equation}

\vspace{-1em}

\begin{equation}
s_{WRC}(d_i)=\frac{\sum^t_{j=1}(\frac{1}{w_j})r_{A_j}(d_i)}{\sum^t_{j=1}\frac{1}{w_j}}
\end{equation}

\vspace{-1em}

\noindent where $w_j\in \{p_j, ds_j\}$, such that $p_j$ and $ds_j$ are the performance and diversity strength of scoring system $A_j$, respectively. 

The performance, $p_j$, is determined by the performance of the scoring system $A_j$, denoted as $p(A_j)$, which is assessed based on performance metrics such as area under the receiver operating characteristic curve (AUROC), accuracy, or precision, depending on the dataset and the specific task at hand. In the current work, we use precision to measure the performance of a model or a combined model. 

In this study, we do not use weighted combination by performance since different human experts may demonstrate different subjectivity for choosing the best SDG label, and we don't have objective ground truth labels. Therefore, we don't use weighting schemes that leverage on performance measures. 

\subsection{Base Models}

In this subsection, we briefly describe the five base models used in our study: A, B, C, D, and E. The first three models, A, B, and C, were designed by domain experts and take into consideration SDG's characteristics. The next two models, D and E, are general machine learning and AI models. 

\subsubsection{SDG Classy (Model A)}

SDG Classy leverages Latent Dirichlet Allocation (LDA), a probabilistic topic modeling algorithm that represents documents as mixtures of topics and topics as distributions over words \cite{jin_natural_2023, lafleur_art_2019}. By allowing users to specify the number of topics, such as the 17 SDGs, LDA minimizes perplexity to learn interpretable topic clusters. Its strength lies in capturing overlaps among goals by assigning words and documents to multiple topics with varying probabilities \cite{blei_probabilistic_2012}, which allows flexible, probabilistic classification and provides semantic insight into SDG-related content.

\subsubsection{LinkedSDG (Model B)}

LinkedSDG (Model B), developed by the UN Department of Economic and Social Affairs (UN DESA) \cite{noauthor_linkedsdg_nodate}, applies semantic web principles to link public content to the SDGs. Using a predefined ontology that formalizes the SDG goal-target-indicator hierarchy, it assigns Internationalized Resource Identifiers (IRIs) to SDG elements which allows precise connections between unstructured documents and SDG concepts. Model B calculates SDG scores by analyzing the frequency of selected concepts and the number of semantic paths linking them to SDG entities, providing structured, interpretable results similar to Model A.

\subsubsection{SDG Mapper (Model C)}

SDG Mapper (Model C), developed by researchers from the European Union, relies on a comprehensive and carefully validated set of SDG-related keywords \cite{s_mapping_2022}. Crafting this keyword set requires interdisciplinary expertise and rigorous cross-verification, as the accuracy and effectiveness of the model depends heavily on the quality of these terms. However, keyword detection in text is not straightforward since variations such as pluralization, conjugation, synonyms, and word order pose challenges that are partially addressed during keyword definition and further mitigated through text preprocessing.

Once the text is preprocessed, it is scanned for keyword matches. Rather than relying on simple counts, SDG Mapper applies a sophisticated aggregation rule that considers text length, keyword frequency, and the number of distinct keywords linked to the same SDG. This nuanced approach helps filter out weak or irrelevant associations so that only goals with sufficient contextual support are considered for final SDG mapping.

\subsubsection{Convolutional Neural Network (Model D)}

Convolutional Neural Networks (CNNs), though designed for images, are effective in text classification by detecting local n-gram patterns through convolutional filters on word embeddings. Pooling reduces dimensionality while retaining key features, which are flattened and fed to dense layers for classification. CNNs are efficient, parallelizable, and parameter-light, making them suitable for large-scale text tasks.

\subsubsection{Random Forest (Model E)}

Random Forest is an ensemble of decision trees built on random data and feature subsets, whose aggregated outputs improve accuracy and reduce overfitting. It handles high-dimensional data with minimal preprocessing, tolerates missing values, and provides feature importance scores.

\section{Experiment Workflow and Datasets Overview}

\begin{figure}
    \centering
    \includegraphics[width=0.7\linewidth]{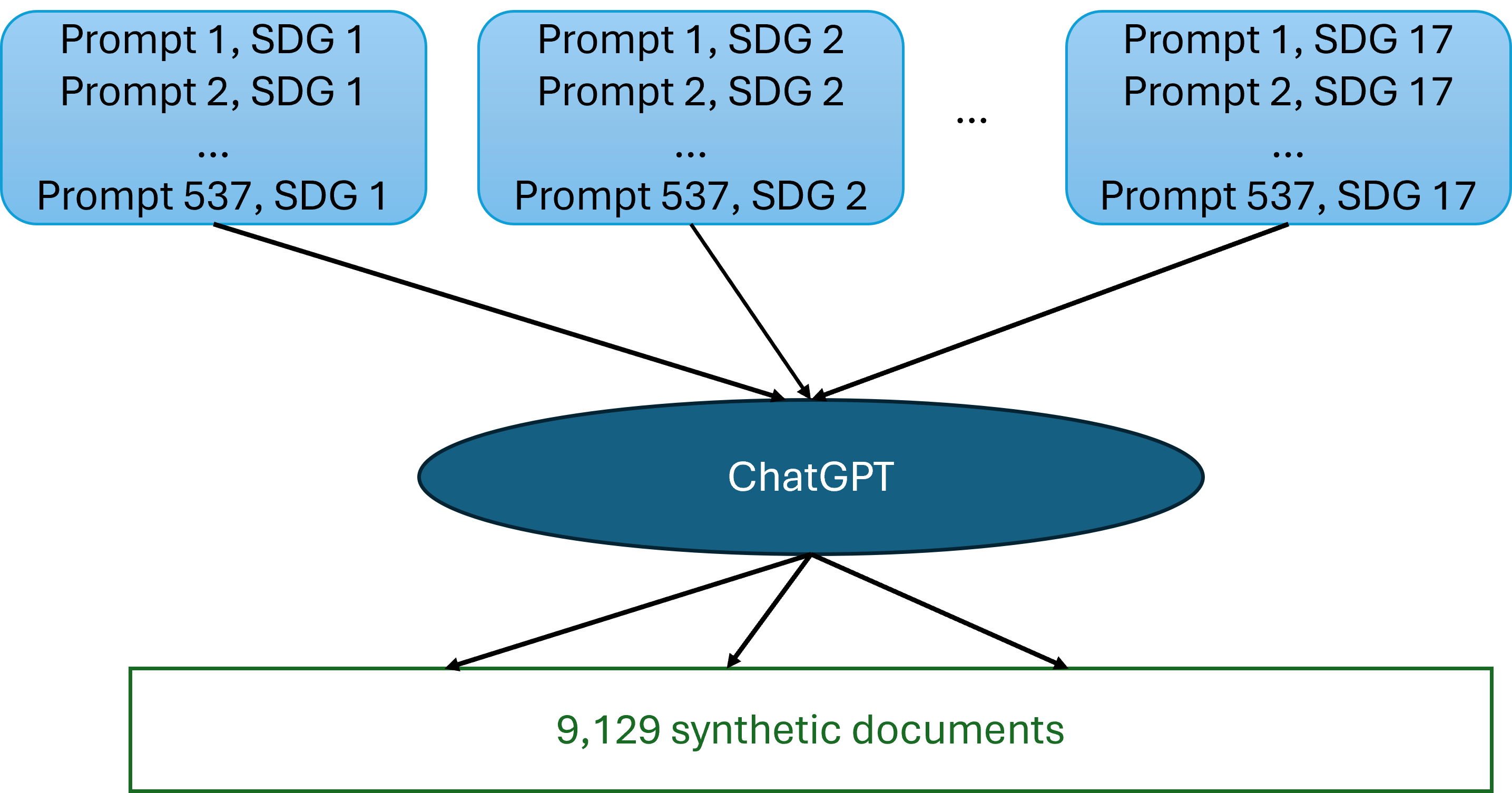}
    \caption{Step 1 of the Methodology Workflow}
    \label{fig:step1}
\end{figure}

\textbf{Step 1:} In addition to three existing SDG-specific models (A, B, and C), we include two conventional models: CNN (Model D) and Random Forest (Model E). To train Models D and E, we use the ChatGPT API to generate synthetic training data based on 537 designed prompts per SDG (Fig. \ref{fig:step1}), styled after 537 unique sources, as shown in Table \ref{tab:variations}. This resulted in 9,129 labeled texts (537 × 17 SDGs). Each text is labeled by the SDG referenced in the prompt. For example, "Write an academic article about SDG 6 by Paul R. Krugman" is labeled as SDG 6. However, due to the interconnected nature of SDGs, each text's label reflects its primary focus, not exclusivity. 

To examine the quality of synthetic data, we evaluated the dataset across three dimensions: document length, lexical diversity (number of unique words divided by total number of words in a document), and phrase diversity (proportion of unique n-grams (e.g., sequences of 2 or 3 words) to the total number of n-grams in the text). The average document length is 628 tokens, and the average type-token ratio (TTR), a lexical diversity measure, is 0.48, which indicates a balanced level of vocabulary variety. Phrase diversity is high, with distinct bigram and trigram ratios averaging 0.88 and 0.96, respectively.

At the per-label level, token lengths are consistent, with the mean ranging from 619 to 632 tokens, and TTR values are stable across goals (0.47-0.50). Bigram diversity remains between 0.87–0.90, and trigram diversity between 0.95 to 0.97. Furthermore, a previous study had also established that this synthetic dataset does well to differentiate UN documents
in 17 SDGs using an SDG-specific model \cite{lafleur_using_nodate}. These results indicate that the prompts we have designed and used have successfully produced balanced outputs for each SDG.

\begin{figure}
    \centering
    \includegraphics[width=0.60\linewidth]{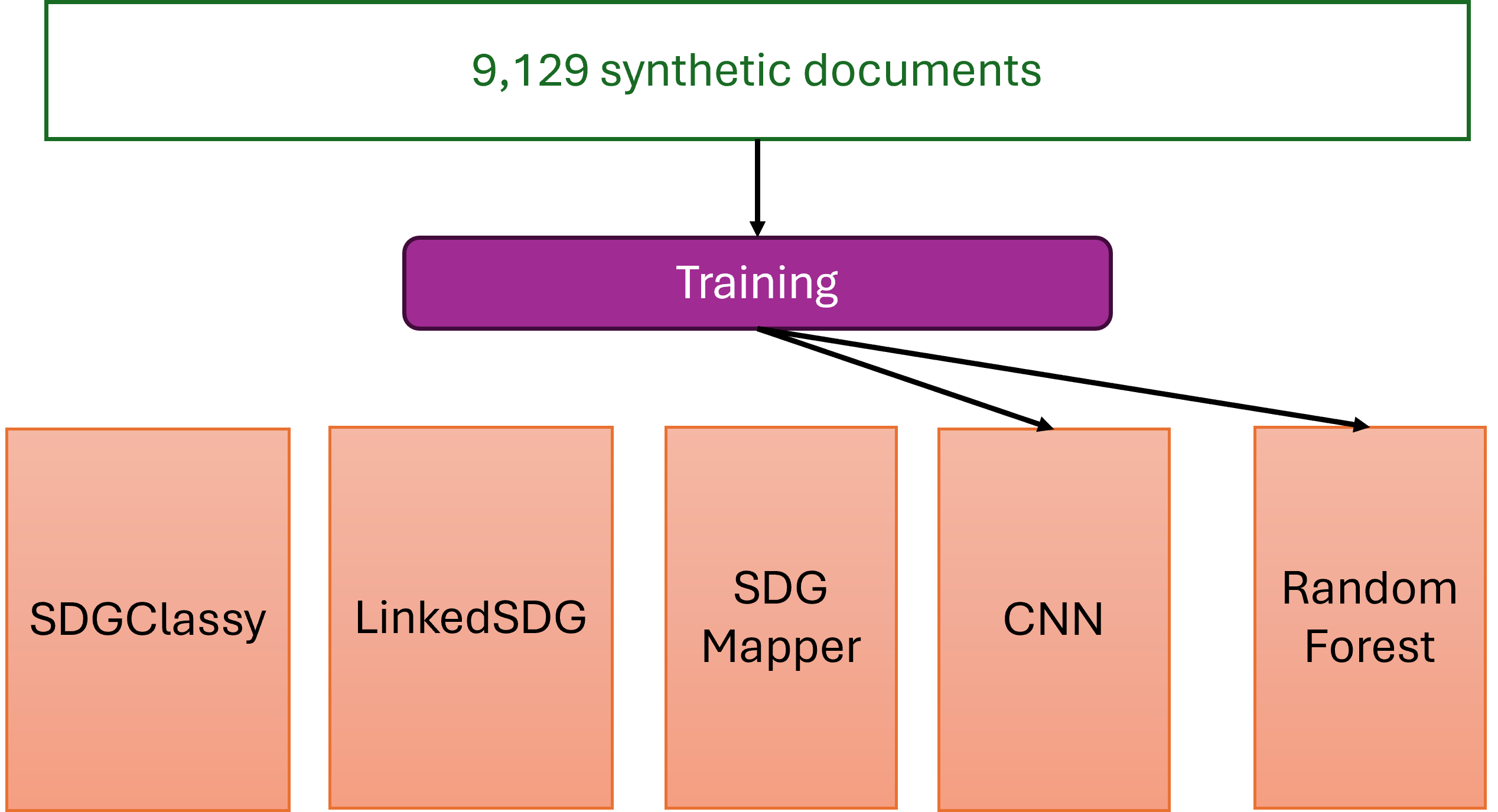}
    \caption{Step 2 of the Methodology Workflow}
    \label{fig:step2}
\end{figure}

\begin{table}[htbp]
    \centering
    \footnotesize
    \setlength{\tabcolsep}{9pt}
    \renewcommand{\arraystretch}{1}
    \begin{tabular}{p{1.1cm}|p{3cm}|p{2.6cm}}
        \hline
        \textbf{Publication Type} & \textbf{Style Variations} & \textbf{Sources} \\
        \hline
        Analytical Reports & UN institutions, departments, and divisions; Various governmental development aid agencies & 
        \begin{tabular}[t]{@{}p{3cm}@{}}
        List of governmental development aid agencies according to Wikipedia
        \end{tabular} \\
        \hline
        Academic Articles & Various scholars & 
        \begin{tabular}[t]{@{}p{3cm}@{}}
        Top 100 economists according to REPEC list
        \end{tabular} \\
        \hline
        Op-eds & Various authors & 
        \begin{tabular}[t]{@{}p{3cm}@{}}
        List of English-language newspaper columnists according to Wikipedia
        \end{tabular} \\
        \hline
        News Articles & Various newspaper and magazines in the US & 
        \begin{tabular}[t]{@{}p{3cm}@{}}
        List of US magazines according to Wikipedia and list of US newspapers according to Wikipedia
        \end{tabular} \\
        \hline
        Academic Journals & Various academic journals & 
        \begin{tabular}[t]{@{}p{3cm}@{}}
        Top 50 academic journals according to REPEC list
        \end{tabular} \\
        \hline
    \end{tabular}
    \caption{Variations in the prompts that generate training data for each of 17 SDGs}
    \label{tab:variations}
\end{table}
\textbf{Step 2:} As we complete the text generation process, we use them to do training, those data are used to pre-train Model D and Model E, as illustrated in Figure \ref{fig:step2}.

\textbf{Step 3:} After Step 2, as each of the 5 models is ready to be used, we apply them on a test dataset which consists of 306 manually curated sample texts, all of which can be found at \url{https://github.com/SeaCelo/SDG-samples/tree/main/datasets/sdg_separates/UN-examples}{https://github.com/SeaCelo/SDG\string-samples/tree/main/datasets/sdg\string_separates/UN\string-examples}. All of these texts are grouped by 17 SDGs, as shown in Table \ref{table:num doc}. 

\begin{table}[H]
\centering
\scriptsize
\begin{tabular}{|| c | c || c | c || c | c ||} 
\hline
SDG & \# of docs & SDG & \# of docs & SDG & \# of docs\\
\hline
SDG 1 & 16 & SDG 7 & 15 & SDG 13 & 13\\ 
\hline
SDG 2 & 21 & SDG 8 & 22 & SDG 14 & 13\\ 
\hline
SDG 3 & 29 & SDG 9 & 22 & SDG 15 & 17\\ 
\hline
SDG 4 & 20 & SDG 10 & 14 & SDG 16 & 16\\ 
\hline
SDG 5 & 23 & SDG 11 & 16 & SDG 17 & 19\\ 
\hline
SDG 6 & 17 & SDG 12 & 13 & Total & 306\\ 
\hline
\hline
\end{tabular}
\caption{Number of documents per SDG in the dataset}
\label{table:num doc}
\end{table}

These manually curated representative texts are gathered from a variety of sources and different time periods, such as text that describes goals, targets, and selected indicators, SDG reports, Secretary General's reports on progress toward SDGs, etc.


It should be noted that each representative text is classified as a single item without any adjustment based on length. Smaller text like indicator descriptions are disproportionately represented in the corpus and is a factor that affects the accuracy in the study results. In addition, the texts were stripped of all figures, footnotes, preambles, tables, bibliographies, and uninformative text in order to be applied to the classifiers. It's to ensure there's a more direct focus on the relevant content. 

\begin{figure}
    \centering
    \includegraphics[width=0.75\linewidth]{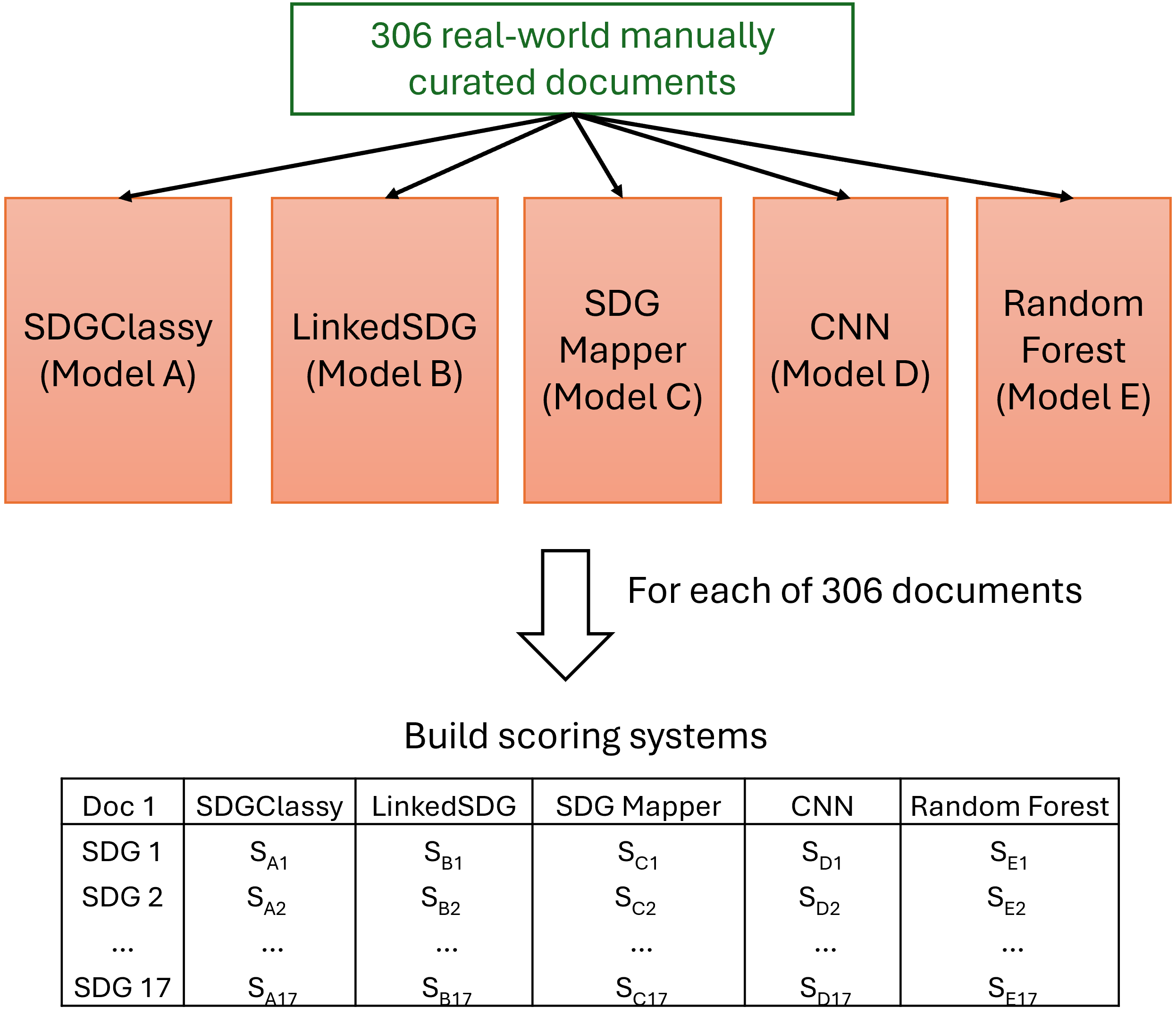}
    \caption{Step 3 of the Methodology Workflow}
    \label{fig:step3}
\end{figure}

Each of the 5 models outputs a score for each SDG for each document being classified, as shown in Figure \ref{fig:step3}. This step establishes 5 distinct scoring systems and we use them to perform model combination in the following step. 

\begin{figure}[H]
    \centering
    \includegraphics[width=0.85\linewidth]{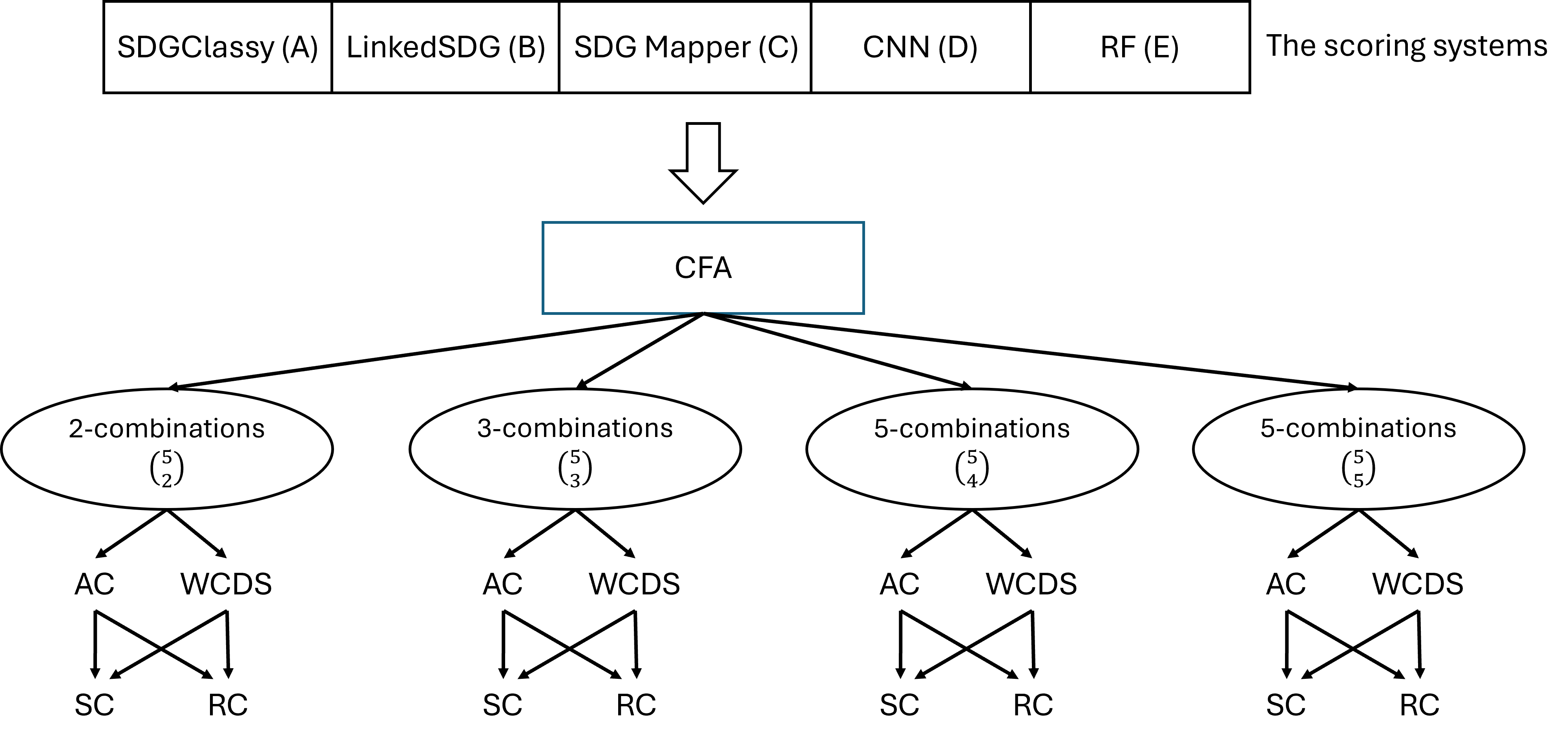}
    \caption{Step 4 of the Methodology Workflow}
    \label{fig:step4}
\end{figure}

\textbf{Step 4}: Figure \ref{fig:step4}, which illustrates Step 4, represents the core function of CFA. The 5 scoring systems are combined by forming various combinations of the five base models, and it creates combination subsets that range from pairs to the full set of five models. All possible combinations involving two, three, four, and five models give us 26 unique model combinations, which is calculated as $5C2+5C3+5C4+5C5=26$.

For each of these 26 combinations, we consider 4 different combination approaches: score combination, rank combination, weighted score combination by diversity strength, and weighted rank combination by diversity strength. As a result, after applying the 4 combination strategies to 26 model combinations, we generate a total of $26\times4=104$ model combinations. 

For each combination, we rank the combination results and compare the top ranked SDGs with human expert opinions to see whether both opinions concur or differ with each other. Our aim is to identify those documents where the experts and machine intelligence opinions differ and seek to gain a better understanding of the interconnection between the SDGs.

\section{Results}
In this section, we discuss the results of applying CFA to the manually curated SDG representative texts by human experts. We then compare the CFA results with those of the five base models as well as the combined models. 

\subsection{Combined Models Results}

We use a measurement, average precision, to understand how much agreement there is between the opinions of human curator and model results. If the results match, then the precision@1, that is precision at rank 1, is 1 for that particular document, because it's 100\% prediction accuracy, and 0 otherwise. For instance, an average precision@1 at 0.9375 for an SDG corpus that consists of 16 documents means there are 15 out of 16 documents where the results of the model and human experts agree with each other.

We document average precision@1 for score and rank combinations, and weighted score and rank combinations by diversity strength for each SDG. The combined models achieved the same or higher average precision than the highest performing individual model in each SDG group 62.89\% (1112/1768) of the time. Furthermore, when we compare the results with the average performance of the 5 individual models for each SDG, the combined models achieve a better result 92.19\% of the time. These results illustrate the robustness of CFA. 

\begin{figure}[H]
    \centering
    \includegraphics[width=0.75\linewidth]{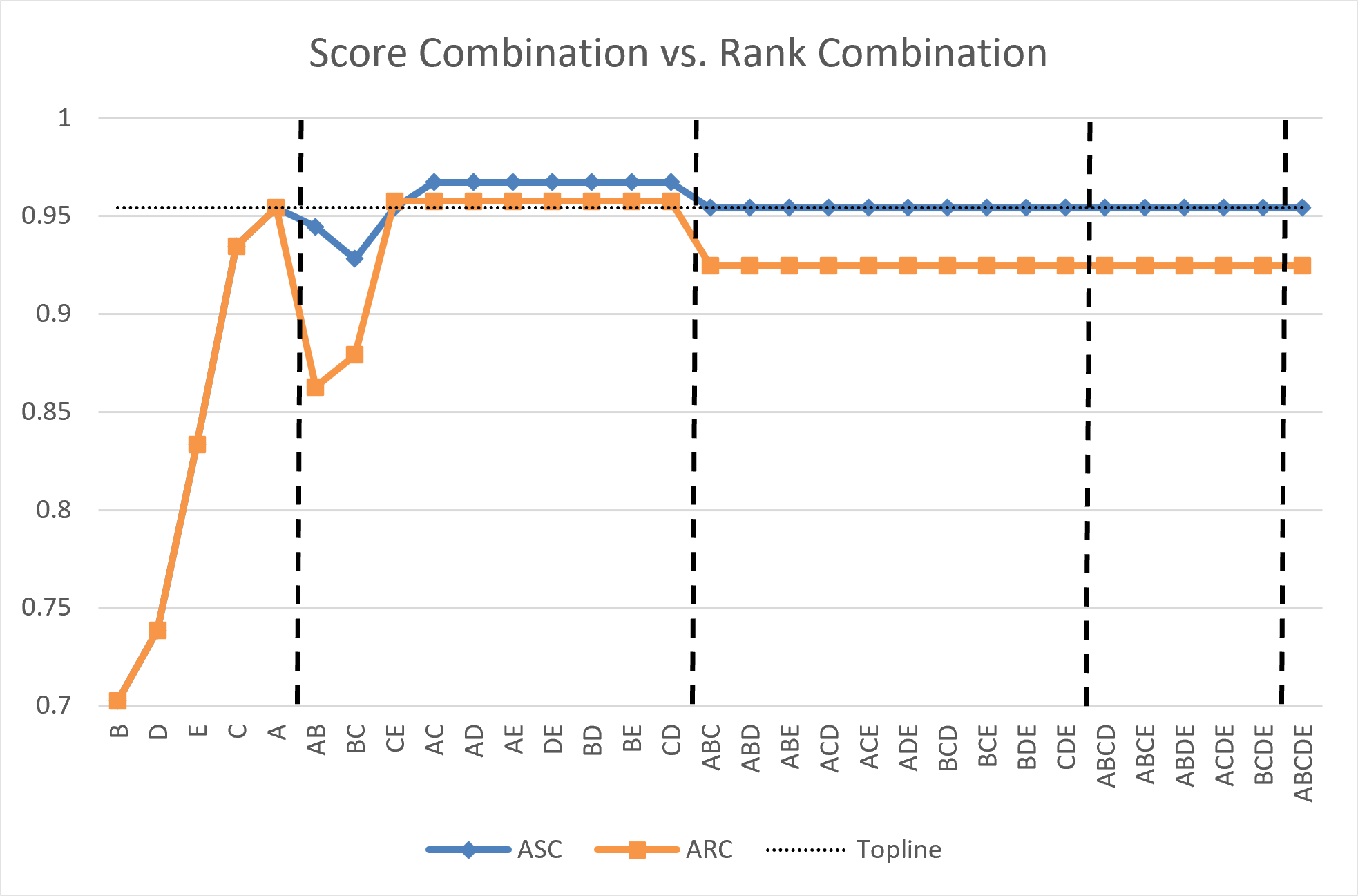}
    \caption{Avg. pre@1 for average combination}
    \label{fig:AC graph}
\end{figure}

\begin{figure}[H]
    \centering
    \includegraphics[width=0.75\linewidth]{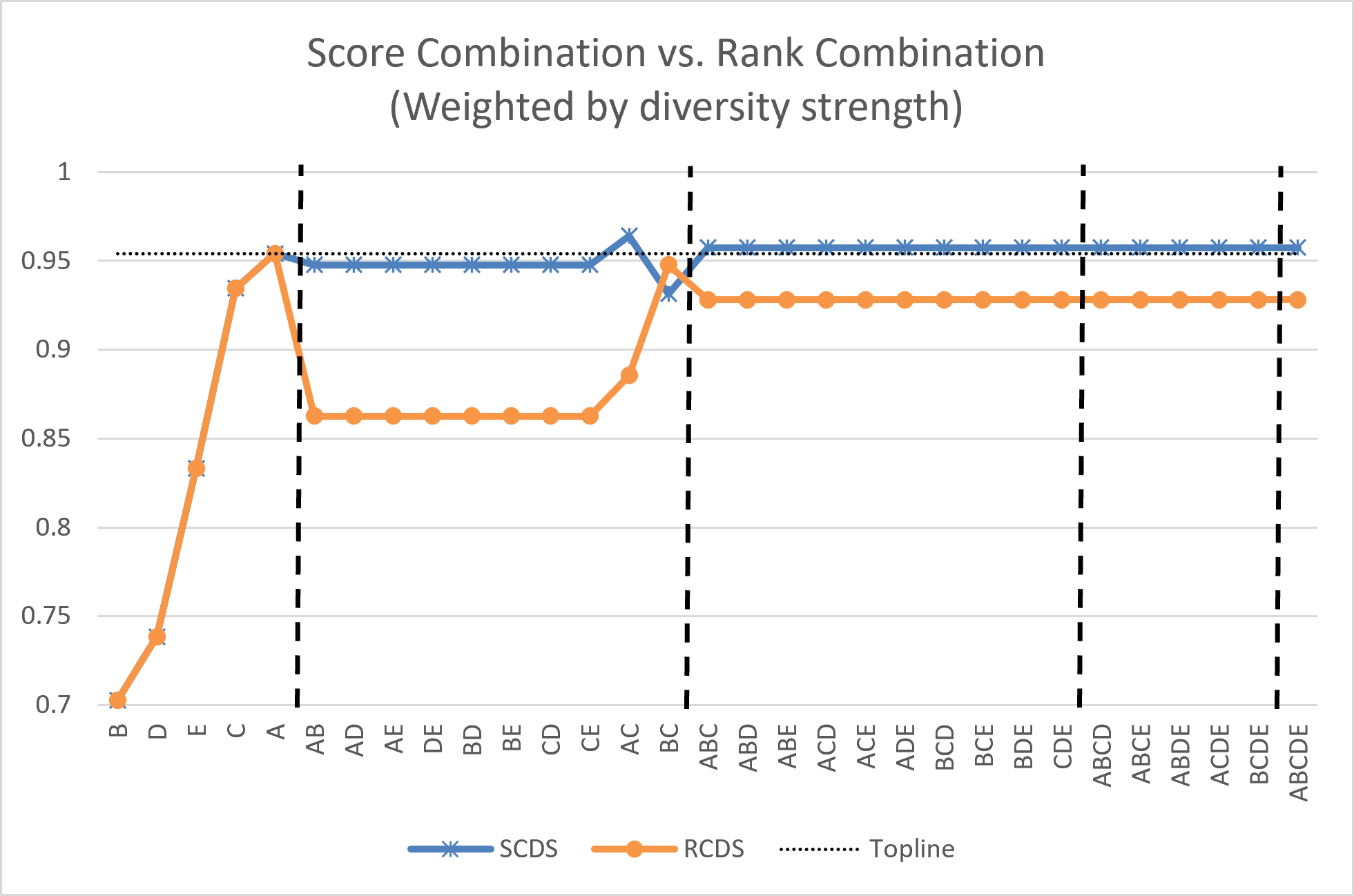}
    \caption{Avg. pre@1 for weighted combination by diversity strength}
    \label{fig:DS graph}
\end{figure}

We also record average precision@1 for score and rank combinations (Figure \ref{fig:AC graph}), and weighted score and rank combinations by diversity strength (Figure \ref{fig:DS graph}) for the entire 306-documents testing set. Model A performs the best among all individual models, having achieved an average precision@1 of 0.9542. We have a number of combined models that outperform Model A, having achieved average precision@1 at \textbf{0.9673}. In general, we have 49, or 47.12\%, of the combined models that either have the same or higher average precision@1 compared to the best individual model.

Further examination of the results shows that when comparing with the best individual model, 3-combination models and weighted score combination by diversity strength are the best in terms of the total number of models that perform better than the best individual model. 

To contextualize CFA performance against a transformer-based model, we fine-tuned a BERT-base model using the same pipeline for training and testing. The BERT model used standard text preprocessing and was fine-tuned for 3 epochs with learning rate 2e-5. BERT achieved an average precision@1 of 0.9446, which is lower than our best CFA combination (0.9673). This demonstrates that while transformer models are strong standalone classifiers, CFA’s model fusion approach can surpass them by leveraging model diversity.

\subsection{Documents for which CFA and Best Base Models Disagrees}

\begin{table}[H]
\centering
\begin{subtable}[t]{0.47\textwidth}
\footnotesize
\centering
\begin{tabular}{|c|p{2.1cm}|p{1.7cm}|p{1.3cm}|p{1.5cm}|}
\hline
\multicolumn{1}{|c|}{} & \multicolumn{4}{c|}{\textbf{Classification Results}} \\ \hline
\textbf{\#} & \textbf{Document} & \textbf{Human Experts} & \textbf{Best Ind. Model} & \textbf{Best CFA Models} \\ \hline\hline
1 & SDG 4 HLPF Thematic Review  & SDG 4  & SDG 12  & SDG 4 \\ \hline
2 & Indicator 5-b-1 & SDG 5 & SDG 9 & SDG 5 \\ \hline
3 & SDG 8 HLPF Thematic Review & SDG 8 & SDG 12 & SDG 8 \\ \hline
4 & Indicator 10-4-1  & SDG 10  & SDG 8  & SDG 10 \\ \hline
5 & SDG 10 HLPF Thematic Review  & SDG 10  & SDG 12 & SDG 10 \\ \hline
6 & SDG 13 HLPF Thematic Review  & SDG 13  & SDG 12  & SDG 13 \\ \hline
7 & Indicator 16-2-3  & SDG 16  & SDG 5  & SDG 16 \\ \hline
8 & SDG 16 HLPF Thematic Review  & SDG 16  & SDG 12  & SDG 16 \\ \hline
\end{tabular}
\caption{CFA agrees with human experts}
\label{tab:typeA}
\end{subtable}
\hfill
\vspace{4mm}
\begin{subtable}[t]{0.47\textwidth}
\centering
\footnotesize
\begin{tabular}{|c|p{1.9cm}|p{1.7cm}|p{1.3cm}|p{1.5cm}|}
\hline
\multicolumn{1}{|c|}{} & \multicolumn{4}{c|}{\textbf{Classification Results}} \\ \hline
\textbf{\#} & \textbf{Document} & \textbf{Human Experts} & \textbf{Best Ind. Model} & \textbf{Best CFA Models} \\ \hline\hline
1 & Indicator 5-a-1   & SDG 5  & SDG 5  & SDG 1  \\ \hline
2 & Indicator 11-6-1  & SDG 11 & SDG 11 & SDG 12 \\ \hline
3 & Indicator 13-1-1  & SDG 13 & SDG 13 & SDG 11 \\ \hline
4 & Indicator 1-5-1   & SDG 1  & SDG 1  & SDG 11  \\ \hline
\end{tabular}
\caption{CFA disagrees with both best base model and human experts}
\label{tab:typeB}
\end{subtable}
\caption{Documents in which CFA and best base model differs}
\end{table}

Out of 306 documents, there are 12 documents for which the best CFA model and the best individual model assign different labels. These documents include long description of indicators and HLPF Thematic Reviews for different SDGs. Out of these documents, there are 8 cases where the best combined models classified the document "correctly" and the best base model was "incorrect", as shown in Table \ref{tab:typeA}. These examples demonstrate the advantage of using CFA as opposed to a single optimized model.

Table \ref{tab:typeB} lists the documents for which the best combined models differ from both the human expert and the best base model. We examine the following samples from this table: Indicator 5-a-1 falls under SDG 5 "Gender Equality" as it measures women ownership of agricultural land. The combined models misclassify this document as SDG 1 "No Poverty," likely since it mentions access to economic resources and financial services. Indicator 11-6-1 monitors urban waste collection and is classified as SDG 11 "Sustainable Cities and Communities." However, the subject of waste collection is also related to SDG 12 "Responsible Consumption and Production." Since indicators can be useful for multiple SDGs, one could consider having multiple class labels for documents for a more comprehensive representation. This would resolve the issue of unfairly marking a prediction as a misclassification.

\subsection{Documents for which CFA and Best Base Models Disagree with human experts}

\begin{table}[h]
\footnotesize
\centering
\begin{tabular}{|c|p{2.1cm}|p{1.5cm}|p{1.5cm}|p{1.5cm}|}
\hline
\multicolumn{1}{|c|}{} & \multicolumn{4}{c|}{\textbf{Classification Results}} \\ \hline
\textbf{\#} & \textbf{Document} & \textbf{Human Experts} & \textbf{Best Ind. Model} & \textbf{Best CFA Models} \\ \hline\hline
1 & Indicator 1-5-2   & SDG 1  & SDG 11  & SDG 11  \\ \hline
2 & Indicator 1-5-3  & SDG 1 & SDG 11 & SDG 11 \\ \hline
3 & Indicator 3-9-2  & SDG 3 & SDG 6 & SDG 6 \\ \hline
4 & Indicator 8-4-2   & SDG 8  & SDG 12  & SDG 12  \\ \hline
5 & Indicator 13-1-2   & SDG 13  & SDG 11  & SDG 11  \\ \hline
6 & Indicator 14-5-1   & SDG 14  & SDG 15  & SDG 15  \\ \hline
\end{tabular}
\caption{Documents where both CFA and best base models disagree with human experts}
\label{tab:type C table}
\end{table}

The documents for which both the best base model and the best combined models differ from the human expert are shown in Table \ref{tab:type C table}. For Indicator 1-5-2, the classification given by human experts is SDG 1, which is fitting since the document discusses building the economic resistance of the poor as a result of disasters. Classifying this document alternatively as SDG 11 which is about "Sustainable cities and communities" is logical as the document discusses infrastructure loss examples such as schools, hospitals, commercial and government buildings, telecommunications, etc. that are integral parts of a city. Indicator 1-5-3 is also related to building the resistance of the poor in times of climate, economic, or social disasters, but this indicator measures countries' national disaster risk strategies. This document is also about SDG 1 and alternatively classified as SDG 11 since it mentions urban resilience and civil protection agencies, which are related to cities. Indicator 13-1-2 is the same measure as Indicator 1-5-3; however, here it is focused on taking urgent action to combat climate change induced natural disaster by strengthening resilience capacity, which is SDG 13 "Climate Action." This was instead classified as SDG 11 by models, since it includes topics such as urban resilience and sustainable development.

Indicator 3-9-2, which measures mortality rate due to unsafe water and sanitation with the goal of promoting health and well-being was classified by the human expert as belonging to SDG 3 "Good Health and Well-Being." Since the document goes into details about unsafe water, drinking water services, and managed drinking water and sanitation services, it overlaps with SDG 6 "Clean Water and Sanitation." 

Indicator 8-4-2 is concerned with material consumption with the goal of improving efficient use of global resources for consumption and production, and is assigned SDG 8 "Decent Work and Economic Growth." It was alternately classified as SDG 12 "Responsible Consumption and Production" since it focuses on consumption in the text, although it could be considered as a means to accomplish SDG 8. 

From these examples, we also see how multi-class labels can address the issue of overlapping concepts in SDGs and indicators.

Indicator 14-5-1 is about protected marine areas and is designated under SDG 14, which focuses on conserving and sustainably developing oceans, seas and marine resources. The document specifies the definition of marine Key Biodiversity Areas and mentions "marine vertebrates, invertebrates, plants, ..." However, a model may focus on the words related to life on land and not capture the marine context, which is why it may have been otherwise classified as SGD 15. This brings attention to the need for context awareness in NLP.

\section{Discussion, Limitation, and Future Work}

In this paper, we apply Combinatorial Fusion Analysis (CFA) to combine five classifier models: A (SDGClassy), B (LinkedSDG), C (SDG Mapper), D (CNN), and E (Random Forest) for document topic modeling aligned with the UN’s 17 Sustainable Development Goals (SDGs). CFA leverages a rank-score function to represent each model and uses cognitive diversity (CD) to quantify their dissimilarity. To address data scarcity, we also employ generative AI tools, specifically ChatGPT, to create synthetic training data.

Our findings show that: (a) several combined models outperform individual ones; (b) diversity among models varies by document and SDG; (c) most models align with expert classifications, though some diverge; and (d) comparative results between combined models and human experts further validate CFA’s effectiveness.

CFA demonstrates how machine intelligence can complement human expertise in AI applications and topic classification. It offers a domain-independent, distribution-free framework that operates in both Euclidean and rank spaces. Its strength lies in mitigating bias and improving generalization through model combination. This work can also easily integrate future UN-affiliated models which makes SDG classification tasks scalable.

While CFA is less prone to overfitting than traditional ensembles, it has limited options for training and tuning. A key limitation is persistent tied rankings, which hinder the effectiveness of diversity-weighted combinations. Future work can explore matrix-based rank representations to address this issue \cite{AKBARI2023102893}.

Another limitation is the number of UN-affiliated base models used. While there are several other existing models in addition to the ones that we discussed in Sec. 2, they are not available for research use due to lack of public APIs. Expanding access to additional SDG tools could significantly enhance performance.

Another possible limitation lies in how we deal with synthetic data generated by large language models. While generative AIs can produce quite realistic training data, the effectiveness of using synthetic data for training is influenced by various factors including the quality, quantity, and diversity of the generated data. The synthetic data generated from LLMs is usually noisy and has a different distribution compared with real-world data, which can potentially have a profound impact on training performance. To mitigate these effects, in the future, we can employ in-context generation techniques, which refers to giving a more specific context to the LLM, such as one-shot and few-shot generation. Even though in this case, our synthetic data quality has been affirmed positively in terms of coverage and spectrum of each SDG, lexical diversity, and phrase diversity, in the event that the quality of synthetic data isn't assured, CFA can help mitigate these risks because of model fusion techniques. While our base models produce a range of performance levels, all our best combined models cover all our base models. In addition, We can also investigate enhancing base model performance by mixing some of the manually curated data we gathered with the synthetic text dataset during training process.

Documents identified in Tables III and IV didn't come from outright misclassification errors but from the inherent interconnectedness of SDGs. In future work, CFA could be extended to multi-label classification by: (a) allowing each model to output a ranked list of top-k SDGs and/or evaluate the performance with metrics such as Hamming loss and micro/macro F1-score. This would more accurately reflect the complex and overlapping nature of SDG policy texts.

Finally, we propose investigating a multi-layer CFA (MCF) framework to further advance classification performance.

\vspace{12pt}

\end{document}